\documentclass{article}

\usepackage[preprint]{neurips_2026}
\workshoptitle{NeurIPS 2026 Workshop on Interpreting Agent Behavior (IAB)}

\usepackage[utf8]{inputenc}
\usepackage[T1]{fontenc}
\usepackage{hyperref}
\usepackage{url}
\usepackage{booktabs}
\usepackage{amsmath,amsfonts,amssymb}
\usepackage{nicefrac}
\usepackage{microtype}
\usepackage{xcolor}
\usepackage{graphicx}
\usepackage{array}
\usepackage{enumitem}

\definecolor{todored}{RGB}{170,30,30}

\newcommand{\stabilityEpisodeRecovery}{92\% (23/25)}
\newcommand{\stabilityGapJSD}{0.0040}

\newcommand{\forecastN}{2{,}975{,}438}
\newcommand{\forecastUsers}{10{,}000}
\newcommand{\forecastFlatF}{0.199}
\newcommand{\forecastFullF}{0.232}
\newcommand{\forecastEpisodeDelta}{+0.023}
\newcommand{\forecastEpisodeCI}{[0.022, 0.023]}

\newcommand{\forecastFullRel}{17\%}
\newcommand{\forecastOpRel}{44\%}
\newcommand{\forecastEpisodeRel}{11\%}

\title{Rhythms of Work: Multi-Scale Interpretation of Human Behavioral Traces for Workplace Agents}

\author{%
  Lin Ai \\
  Microsoft \\
  \texttt{lai@microsoft.com}
  \And
  Scott Counts \\
  Microsoft \\
  \texttt{counts@microsoft.com}
}

\begin{document}
\maketitle

\begin{abstract}
% \scott{feel free to leave inline comments like this} 
Runtime traces are becoming a central substrate for understanding agentic systems, yet interpretation has focused largely on what the agent did. Workplace agents face the complementary problem: interpreting the human activity that surrounds them. Hours of low-level events carry rich evidence about a user's state but are too granular to reason over directly, and flattening them into one stream or compressing them into a single embedding both treat ``summarize the user's behavior'' as if it had one correct answer. We argue instead that behavioral interpretation is resolution-dependent: the same trace should admit multiple addressable interpretations at different temporal resolutions. We construct a multi-resolution vocabulary of semantically normalized operators, recurring motifs, coherent episodes, and day-level rhythms, each preserving the structure salient at its own horizon. Applied to 667 million human-attributed events from a large commercial productivity suite (50,000 users, 100 organizations), it yields 120 operator types, thousands of motifs, 25 episode types, and five day-rhythm archetypes. We validate it on real telemetry: re-running the entire pipeline on a disjoint 2,000-user sample recovers the same taxonomy (structural stability), and on held-out users the full representation forecasts a user's next episode more accurately than a flat-operator baseline, a \forecastFullRel{} relative macro-F1 gain (predictive validity), so the abstractions preserve future-relevant information rather than merely describe it. A controlled resolution ablation then shows that no single level is optimal across questions: different agent-facing questions about the same trace are best answered at different resolutions. Behavioral trace interpretation for agents should therefore be multi-resolution and query-conditioned: an agent should access the temporal grain a question needs, not one universal summary.
\end{abstract}

\section{Introduction}

Agentic systems increasingly produce and consume long behavioral traces, and a growing body of work asks how to interpret them \citep{gao2026interpret, ou2025agentdiagnose}, most of it looking at the agent's own trajectory: what the model did and why. Workplace assistants face a complementary problem that has received far less attention, making sense of the \emph{human} trajectory surrounding them. Before, during, and between interactions, a person leaves a dense stream of edits, messages, file changes, application switches, and meetings, richly informative about how their work is unfolding but, in raw form, not an interpretable account of their behavior.

The difficulty is that the trace is both too long and too low-level to reason over directly, and the usual ways of handling it discard exactly what an agent needs. A flat event sequence preserves every local detail but offers no explicit interpretation, forcing each downstream model to reconstruct longer-range structure from scratch. A single summary or embedding is compact but hides which temporal horizon the interpretation rests on. Both treat ``summarize the user's behavior'' as if it had one correct answer.

\begin{figure}[t]
    \centering
    \includegraphics[width=0.82\linewidth]{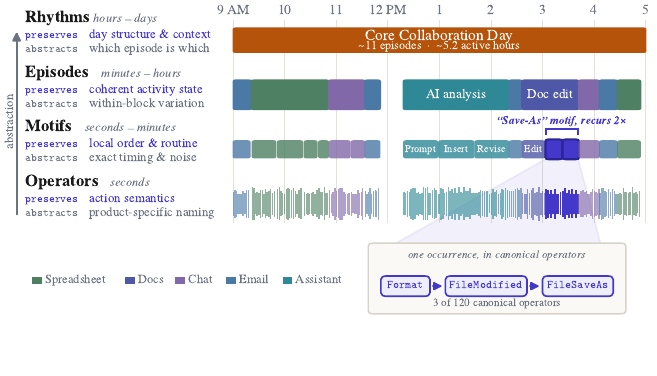}
    \caption{The same workday admits four simultaneous interpretations at different temporal resolutions, read here on one shared clock. The whole day is a single \emph{rhythm}; it is a sequence of \emph{episodes}; each episode is a stream of \emph{operators}; and short recurring sub-sequences of operators are \emph{motifs}. Because every track shares the same time axis, containment is visible directly: an episode is a span of operators, a motif a shorter recurring span. The left column states each level's invariance: the time horizon it covers, what it preserves, and what it abstracts away, which is what lets the levels be addressed separately rather than fused into one summary. The highlighted motif is a word-processor ``Save-As Formatting Pass'' that recurs twice in the afternoon; both occurrences normalize to the same operator sequence, Format$\rightarrow$FileModified$\rightarrow$FileSaveAs (3 of 120 canonical operators). The four levels are a conceptual multi-resolution vocabulary; the learning pipeline need not implement a strictly sequential generative hierarchy.}
    \label{fig:representation}
\end{figure}

We argue instead that behavioral interpretation is inherently \emph{resolution-dependent}. The same underlying activity can be read at several valid scales, and each answers a different question: an \emph{operator} names what action just happened, a \emph{motif} what local routine is unfolding, an \emph{episode} what coherent mode of work the user is in, and a \emph{rhythm} what longer pattern of the day contains the current activity. These are not competing labels for one thing; they are addressable views of one trace, each preserving the structure salient at its horizon and abstracting the rest away (Figure~\ref{fig:representation}).

The problem, then, is not only to compress behavior into a summary, but to expose the right interpretation for the question being asked. An agent deciding whether now is a good moment to help, what a user is working on, or whether a recurring pattern is worth acting on is really asking about different temporal grains of the same trace. This reframes representation design around \emph{addressability}: keeping the levels separately retrievable rather than fusing them into one stream or one embedding.

We build such a representation from real workplace telemetry and ask three questions of it. Does the vocabulary re-emerge on behavior it has never seen (structural validity)? Does the interpretation preserve information about how behavior actually unfolds, or is it a readable but inert summary (predictive validity)? And does the useful interpretation depend on the question, or would a single level do (resolution localization)? These three threads organize the paper.

We make three contributions:
\begin{enumerate}[nosep]
    \item \textbf{A multi-resolution vocabulary for interpreting human behavioral traces.} We turn 667M human-attributed workplace events into semantically normalized operators, recurring motifs, coherent episodes, and day-scale rhythms, each with an explicit temporal horizon and stated invariance (120 operators, thousands of motifs, 25 episode types, five rhythms).
    \item \textbf{Evidence that the interpretations preserve real behavioral structure.} On real telemetry the vocabulary replicates on a disjoint 2,000-user re-run and improves next-episode forecasting on held-out users by \forecastFullRel{} relative macro-F1 over a flat-operator baseline, so the abstractions are reproducible and predictive, not merely descriptive.
    \item \textbf{A characterization of which interpretation an agent needs.} Through a controlled resolution ablation, we show that different questions about the same trace are best answered at different temporal grains, and that indiscriminately concatenating all levels can hurt, so the levels should stay addressable rather than fused.
\end{enumerate}

Two scope notes. First, our telemetry captures the full arc of a user's workday rather than only the sessions in which they interact with an agent, so we study the broader human-side behavioral context an agent may need to interpret before, during, or between interactions. Second, we study what an agent should read from a user's behavior, the state layer that any decision to act would build on. Turning that into a policy for whether and how to help also weighs the user's latent need, the cost of interrupting, consent, and the counterfactual, and is a natural next step our representation is meant to enable.

\section{Related work}

\paragraph{Interpreting agent behavior.}
A growing line of work interprets what agents themselves do at runtime, building taxonomies of agent actions from execution traces \citep{gao2026interpret} and toolkits that diagnose agent trajectories by quantifying competencies and visualizing their semantics \citep{ou2025agentdiagnose}. A parallel line of benchmarks asks whether an agent can finish tasks across real applications, on the open web \citep{deng2023mind2web, zhou2024webarena}, across an operating system \citep{xie2024osworld, pmlr-v267-bonatti25a}, and in office suites \citep{pmlr-v235-drouin24a, boisvert2024workarena++, wang2024officebench, wang2025odysseybench}, scoring whether it reaches a goal. Both look at the agent side and say little about the \emph{user} beside the agent. We study the complementary human side: how an agent should interpret the longitudinal human trace around it, when the same underlying activity admits different valid descriptions at different temporal resolutions. Rather than one universal behavioral label, we construct separately addressable operator-, motif-, episode-, and rhythm-level views and test both whether they preserve real structure and which views different questions need.

\paragraph{Behavioral trace mining and process discovery.}
Turning raw event logs into structure is an old goal: sequential pattern mining finds frequent sub-sequences \citep{pel2001prefixspan}, and process mining discovers process models and cases from logs \citep{van2016data}, with object-centric variants relaxing the one-case-per-event assumption \citep{van2019object}. Our motifs and episodes are closely related to these units. Our added claim is that no single discovered unit is a sufficient interpretation on its own: each level carries a stated invariance, and several temporal resolutions should coexist and stay separately addressable rather than collapsing into one flat log or a single notion of a case.

\paragraph{Predictive user modeling.}
Predicting the next event tests whether a representation preserves future-relevant structure: predictive process monitoring forecasts the next activity in a running case \citep{tax2017predictive}, next-action prediction forecasts a user's upcoming action from interaction history \citep{shaikh2026learning}, and app-usage models predict the next application opened \citep{zhao2019appusage2vec}. We use next-episode forecasting in this spirit, as an external validity test with the model class held fixed, not as a forecasting system: does adding a coarser scale preserve information about what happens next?

\paragraph{Proactive agents and user simulation.}
A parallel line builds agents that act before being asked, inferring intent from activity and sensor streams \citep{lu2025proactive, yang2026contextagent}, timing assistance against the user's working memory \citep{pu2025promemassist}, and evaluating with tool use and simulated users \citep{yao2024taubench, yoon2024evaluating}; interrupting at a coarse task boundary costs far less than mid-step \citep{adamczyk2004if}. Our work sits one layer below: intervention policies need behavioral state, but representation is not intervention, and we use synthetic timelines only as instrumented test fixtures, never as faithful user simulation.

\section{A multi-resolution vocabulary for human behavioral traces}

We seek not one summary of a behavioral trace but a set of addressable interpretations at different temporal horizons, each answering a different question about the same events. An \emph{operator} answers what action just happened, preserving action semantics while abstracting away product-specific naming; a \emph{motif} answers what local routine is unfolding, preserving local order while abstracting exact timing and interface noise; an \emph{episode} answers what coherent mode of work the user is in, preserving activity state while abstracting within-block variation; and a \emph{rhythm} answers what longer pattern of the day contains the activity, preserving cross-episode day structure while abstracting exact episode instances. Figure~\ref{fig:representation} states each contract next to one representative day read at all four grains at once. The vocabulary is \emph{multi-resolution}, but the mining pipeline is not a strict bottom-up chain: we segment episodes before mining motifs, because motif vocabularies learned within a cluster of similar activity are cleaner than motifs pooled across all behavior. The level ordering describes what each grain covers, not the order in which the estimators are fit.

We learn this vocabulary from enterprise productivity-suite telemetry: 667 million human-attributed events from 50,000 users across 100 organizations over two months, spanning 1.82 million user-days. Each event carries only an action name, a timestamp, and its originating application. None of this structure is labeled: with no task labels, episode boundaries, workstream identifiers, or personas to fit to, every level in Table~\ref{tab:data} is discovered unsupervised and data-driven rather than annotated.

\begin{table}[t]
\centering
\small
\caption{Scale of the corpus and learned representation. All counts refer to the human-attributed event stream.}
\begin{tabular}{lrrl}
\toprule
Level & Instances / coverage & Vocabulary & Typical scale \\
\midrule
Raw events (human) & 667M events & $\sim$14K app--action pairs & sub-second to minutes \\
Canonical operators & 667M mapped events & 120 types & seconds \\
Motifs & 20.6M occurrences & 3{,}597 motifs & seconds to minutes \\
Episodes & 16.7M instances & 25 types & minutes to hours \\
Day rhythms & 1.82M user-days & 5 archetypes & hours to days \\
\bottomrule
\end{tabular}
\label{tab:data}
\end{table}

\subsection{Canonical operators}

Raw telemetry names actions in each product's own vocabulary: the six core applications emit a long tail of roughly 14,000 distinct $(\text{app},\text{action})$ pairs that mix genuine user actions with interface primitives and machine-generated events. To recover the \emph{meaning} of an action independently of product naming, we map these pairs onto a fixed catalog of canonical operators in two LLM-assisted passes, both using GPT-5.4. The catalog is built primarily from a production audit-log schema,\footnote{Public audit-log activity catalog; source URL withheld for anonymity.} extended with semantic operators for common actions the schema does not name (archiving a message, previewing an attachment) and a few interface primitives (\texttt{Edit}, \texttt{Format}, \texttt{Navigate}, \texttt{Select}) for in-application interactions that are not audit events. The first pass assigns an operator wherever one fits; the second resolves the remainder or discards events that appear non-human. Operators below our privacy threshold are pooled into a single \texttt{Other} category, leaving a final vocabulary of 120 canonical operators.

\subsection{Episodes and motifs}

We segment each user's stream at the gaps between events. Gap lengths span several orders of magnitude, so a fixed threshold either chops long blocks in half or fuses unrelated activity; instead we fit a Gaussian mixture to the log gaps and cut when the posterior of the longest-break component clears 0.8 \citep{mclachlan2019finite} (characteristic scales near 2, 8, 55, and 362 seconds). Each session becomes a 46-dimensional feature vector capturing application fractions, action mix, meeting, call, and AI-assistant context (the fraction of events during a meeting, during a call, or near an AI-assistant interaction), timing, and diversity. Because work modes differ sharply across applications, we cluster sessions within each dominant application using $k$-means with $K\in\{3,4,5\}$ by silhouette score \citep{rousseeuw1987silhouettes}, then merge adjacent same-cluster sessions within 30 minutes so a briefly interrupted block is recovered as one episode. This gives 25 episode types.

Inside each episode cluster we mine contiguous two- to six-token $n$-grams that clear an adaptive support threshold of $\max(20,0.05\times\text{cluster size})$, which prevents a motif from being declared on a few coincidental repeats in a large cluster; mining within a cluster rather than globally preserves local routines that a pooled vocabulary would obscure. Over both raw $(\text{app},\text{action})$ tokens and canonical operators this gives 3{,}597 released motif patterns across 20.6 million occurrences. GPT-5.4-generated labels are attached only after mining and are never optimization targets.

\subsection{Rhythms and population variation}

To describe the shape of a day, we build for each user-day a $24\times25$ hour-by-episode matrix, compress it with truncated SVD to its dominant daily patterns, and append a few scalar summaries (episode count, active duration, meeting and AI-assistant share, and episode entropy). Clustering these with $k$-means gives five day-rhythm archetypes (Figure~\ref{fig:dayrhythms}, Appendix~\ref{app:rhythms}).

Finally, a second clustering over 37-dimensional per-user summaries gives ten personas (Figure~\ref{fig:dayrhythms}, bottom), which we read as \emph{dialects}: distinct long-run mixtures over the same rhythms and operators rather than separate vocabularies. They matter here only as evidence that a single multi-resolution vocabulary describes a heterogeneous population, not as a taxonomy we rely on.

\section{Do the behavioral interpretations capture real structure?}

The previous section produced a four-level vocabulary, but building it does not make it trustworthy: a discovery pipeline like this can overfit its own sample, or replicate perfectly yet carry no information about how work unfolds. We test each failure separately. \emph{Structural validity} asks whether the vocabulary re-emerges on data it has never seen; \emph{predictive validity} asks whether the interpretation preserves information about what a user does next.

\subsection{Structural validity: independent held-out replication}

This first experiment targets the overfitting failure mode: a vocabulary is only useful if it is not an artifact of one sample. If the structure is a genuine property of the population, re-running the pipeline on a fresh sample of users should recover the same gap scales and episode types; if it is an artifact of one clustering run, it should not. We therefore re-run the entire pipeline on a separate, disjoint 2{,}000-user sample and compare it against the full 50{,}000-user run. The gap model's mixture parameters are not directly identifiable, so we compare the fitted densities with Jensen--Shannon divergence \citep{lin1991divergence}. Because each run selects its own number of per-application episode clusters, we match the two cluster sets by Hungarian assignment \citep{kuhn1955hungarian} on the cosine similarity between their operator-frequency profiles.

The gap densities agree to a Jensen--Shannon divergence of \stabilityGapJSD{}, well under our $0.05$ threshold, and \stabilityEpisodeRecovery{} of the 25 episode types are recovered under matching. A same-scale split-half of the 2{,}000-user sample gives the same result (gap-density divergence $0.021$), so the structure is stable across scale and across disjoint samples (density overlay and per-application matches in Figure~\ref{fig:stability}, Appendix~\ref{app:stability}).

\subsection{Predictive validity: focused next-episode forecasting}

Replication rules out the first failure mode but not the second: a vocabulary can be perfectly stable and still discard the information that matters. An interpretation should compress the trace without throwing away what is relevant to how behavior unfolds, so we use forecasting as an external validity test: if the abstractions preserve future-relevant structure, they should make the near future more predictable, and the target can be read directly from telemetry without extra labeling. Our aim is not to build the best possible forecaster but to measure what each level \emph{adds}. We therefore hold two things fixed and vary only the interpretation: a single linear probe, so any difference reflects the features rather than model capacity, and a single coarse target, so the comparison stays clean. We add the levels one at a time, from a near-trivial baseline up to the full stack, and read off the incremental gain at each step. This trades predictive ceiling for interpretability: it cannot tell us the best achievable accuracy, but it can tell us whether each added scale carries real signal.

\begin{figure}[t]
    \centering
    \includegraphics[width=0.80\linewidth]{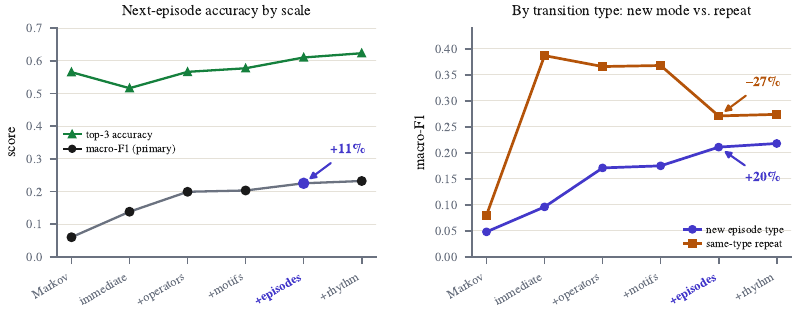}
    \caption{Next-episode forecasting on held-out users (\forecastN{} transitions, \forecastUsers{} users), one fixed linear probe per condition. \textbf{Left:} both metrics rise as representation scales are added cumulatively (macro-F1 primary, top-3 accuracy secondary); adding episodes lifts macro-F1 by \forecastEpisodeRel{} (the pre-registered primary; absolute \forecastEpisodeDelta{}, 95\% CI \forecastEpisodeCI{}). \textbf{Right:} the same conditions split by what happens next. Adding episode context improves new-mode prediction ($+20\%$ relative) but degrades same-type repeats ($-27\%$), where the current type already predicts the next. The average gain is real forward-looking signal, and added scale is not uniformly useful, which is the predictive-side motivation for scale-selective retrieval.}
    \label{fig:forecast}
\end{figure}

We forecast a single target, the type of the user's next \emph{episode}. Of the candidate targets it is the one that genuinely exercises a multi-scale interpretation: the next action or motif is dominated by the finest grain, and a day-rhythm is only defined once the day is over, so supplying it mid-day would leak the future. The episode sits at the useful middle, a coherent unit an agent reasons about, with an objective label, whose prediction can draw on every level at once, from recent operators and motifs to the current episode and the day so far. Appendix~\ref{app:target} gives the full argument for ruling out the other targets.

For every completed episode $e_i$ followed by another later the same day, the model sees the history $H_i=\{e_1,\ldots,e_i\}$ and predicts the type of the next episode $e_{i+1}$, a 25-way classification at each boundary. Every condition uses the same L2-regularized logistic-regression classifier (SGD, balanced class weights), so any difference reflects the interpretation rather than model capacity. The conditions accumulate one level at a time:
\begin{enumerate}[nosep]
    \item \textbf{Markov baseline:} empirical $P(e_{i+1}\mid e_i)$ from the current episode type.
    \item \textbf{Immediate context:} time of day, day of week, current application, and the latest operator.
    \item \textbf{+ operators:} canonical-operator profiles over a same-day episode window (current and previous two individually, then 3rd--5th and 6th--10th bands).
    \item \textbf{+ motifs:} the motif distribution over that window.
    \item \textbf{+ episodes:} episode types, durations, gaps, and type-to-type transitions over the same window.
    \item \textbf{+ rhythm-so-far:} the day-so-far prefix (cumulative active time, episode mixture and entropy, meeting and AI-assistant share, hour-of-day).
\end{enumerate}
Every level uses the same-day prefix only, nothing after the boundary and nothing from earlier days; the last condition never sees the final full-day rhythm label. We split users 70/10/20 by organization, and fit the motif vocabulary, episode centroids, SVD projections, and normalization statistics on training users, then freeze them for the held-out ones. Macro-F1 is primary, top-3 accuracy secondary, with user-level bootstrap confidence intervals. We fixed the confirmatory comparison before seeing the test set, the \emph{episode increment} (the gain from episode-level structure on top of operators and motifs), and emphasize the full representation against a flat-operator baseline.

On \forecastN{} held-out same-day transitions from \forecastUsers{} users, the full representation improved next-episode macro-F1 from \forecastFlatF{} (flat operators) to \forecastFullF{}, a \forecastFullRel{} relative gain, and top-3 accuracy from $0.566$ to $0.623$ ($+10\%$); a Markov baseline reached only $0.060$ (Figure~\ref{fig:forecast}, left). Each scale added signal in the expected order, every per-layer increment with a 95\% bootstrap CI excluding zero. Canonical operators produced the largest single jump ($\forecastOpRel{}$ relative, $0.138\to0.199$); the episode increment, the pre-registered confirmatory comparison, was next at $\forecastEpisodeRel{}$ (absolute \forecastEpisodeDelta{}; 95\% CI \forecastEpisodeCI{}; top-3 $+6\%$). Motifs added $+2\%$ and whole-day rhythm-so-far $+3\%$.

As expected, the full representation outperforms flat operators on held-out users, and the gains are uneven. Operators carry most of the signal, episodes add the largest jump from a representation level, and rhythm-so-far adds only marginally. The key point is that more scale is not always better: episode context helps when the user is about to switch modes but degrades prediction when the user simply continues the same activity (Figure~\ref{fig:forecast}, right; full breakdown in Appendix~\ref{app:forecastsensitivity}). The representation is thus stable and informative, but this last result already hints that which level an agent should read depends on the question, which we turn to next.

\section{Which behavioral interpretation does an agent need?}

The vocabulary is stable and predictive, but both are properties of the interpretation in isolation. Given several valid interpretations of one trace, which should an agent use for a given question? We hold the moment, the question, and the answering model fixed, vary only which level the agent is shown, and measure how well each answers each question. This isolates where the information sits and whether any single interpretation suffices, measuring which grain carries the answer rather than task success.

We probe seven questions an agent might ask about a user's state, in three families. \emph{\textbf{Perception}} (what is happening now?): the current micro-routine at the motif scale, the activity's aim at the episode scale, and the shape of the day so far. \emph{\textbf{Inference}} (what is coming or hidden?): whether the user will stay or switch within the next half hour, and whether attention is fragmented across workstreams. \emph{\textbf{Action}} (what should the agent do?): how disruptive it would be to interrupt now, and which kind of proactive help fits a freshly opened task.

The experiment needs ground-truth labels at every scale, which real telemetry does not provide; and since the representation is content-agnostic, using only operators and their timing, a synthesized trajectory reproduces everything it reads. We therefore run it on \emph{instrumented test fixtures}: top-down synthesized timelines whose latent labels are known by construction and whose surface statistics are calibrated to the real stream, letting us remove one interpretation layer at a time. They are validated for coherence, behavioral fidelity, and timing against the real data but remain synthetic rather than real people, so they let us pin down \emph{where information sits under controlled conditions}, not how often such moments arise in deployment. The full construction, its validation, and the eight context shapes appear in Appendix~\ref{app:synthetic}.

We instantiate the seven capabilities into 640 concrete questions, sampling diagnostic moments across the fixtures' persona timelines where the answer hinges on a specific grain (per-capability counts in Appendix~\ref{app:synthetic}). Each is rendered into eight context shapes, each exposing a different interpretation of the same moment. All share a common baseline (persona role, time of day, and the open artifact), matching what a content-driven agent sees today. Four singletons add one observable layer each (\texttt{+raw}, \texttt{+motif}, \texttt{+episode}, \texttt{+rhythm}); \texttt{+hierarchy} stacks the three abstracted layers (the pipeline's output); \texttt{+oracle} substitutes declared metadata telemetry cannot observe (workstream identity and workflow goal); and \texttt{all} combines them. Every (question $\times$ shape) pair is answered by GPT-5.4 under a rule-out protocol, naming a plausible alternative and explaining why the evidence excludes it, over three runs at temperature zero. We report macro-F1 across runs; run-to-run standard deviation stays under 0.055 for every (shape $\times$ capability) pair.

\subsection{Where each capability finds its answer}

\begin{figure}[t]
    \centering
    \includegraphics[width=0.80\linewidth]{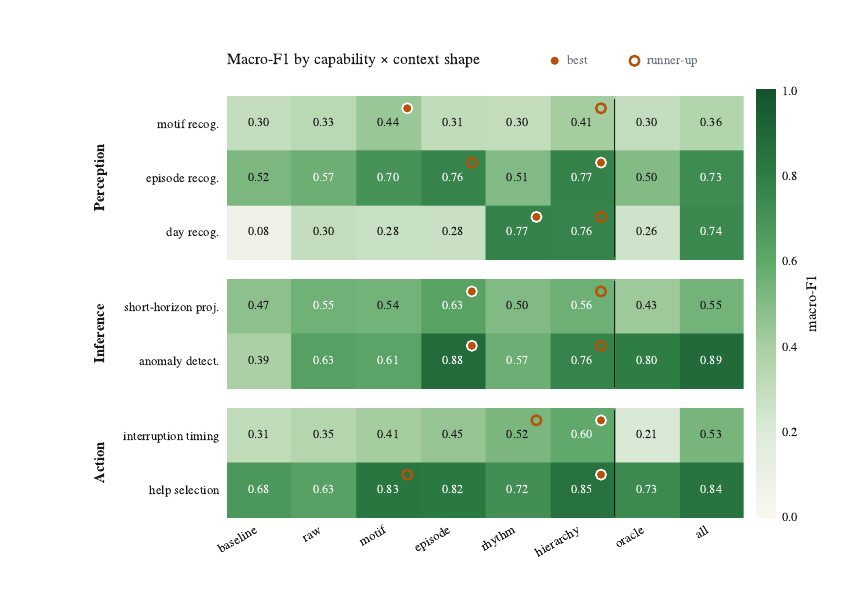}
    \caption{Which interpretation answers each question? Macro-F1 for seven capabilities across eight context shapes, on the instrumented fixtures (640 questions, three runs); each column exposes a different interpretation of the same underlying trace. Rows are grouped into the three families (Perception, Inference, Action). In each row a filled dot marks the best observable context and an open dot its runner-up, so the distance between them shows how sharply the winning grain leads. The best cell moves across four different columns (motif, episode, rhythm, and the stacked hierarchy), so no single interpretation wins everywhere. The two columns to the right of the rule, \texttt{oracle} and \texttt{all}, carry declared metadata that telemetry cannot observe.}
    \label{fig:scorecard}
\end{figure}

\textbf{No single interpretation resolution is universally sufficient.} Different questions are answered best at different levels (Figure~\ref{fig:scorecard}). Across the six observable layers the best cell moves through four of them: the motif layer for motif-scale recognition, the episode layer for behavior projection and anomaly detection, the day rhythm for day-scale recognition, and the full hierarchy for episode-scale recognition, interruption timing, and help selection. Averaged over the seven capabilities the hierarchy leads at 0.67 macro-F1, but only narrowly over the strictly larger \texttt{all} shape and by a few points over the best single grain, and that average hides the real point: the level that is best for one question is not best for another, and adding more layers can lower accuracy rather than raise it. The questions fall into three groups, which we take in turn: those answered at a fine grain, those answered only at a coarse grain, and those that need several grains at once.

\textbf{Fine grains answer questions grounded in the action sequence.} Motif-scale recognition, identifying which micro-routine the user is in, turns on the ordering of small actions: only the motif layer clears the baseline (0.44 against 0.30), while an episode label keeps the topic but discards the action pattern and a day rhythm discards it entirely. Behavior projection, predicting whether the user will stay on task or switch within the next half hour, loads on the same recent grain, and here the coarser layers do positive harm: adding the day rhythm injects a misleading prior that pulls the stacked hierarchy (0.56) below the episode layer that answers it best (0.63).

\textbf{Coarse grains answer questions that exist only at a longer horizon.} Day-scale recognition, whether the day reads as deep work, a full workday, or a light day, describes categories with no meaning below the hour scale, so the day rhythm supplies almost all of the signal, lifting accuracy from 0.08 at the baseline to 0.77, the largest single-layer gain in the study. Episode-scale recognition and anomaly detection sit one level finer and are both answered best at the episode grain, where the observable episode layer already reaches the ceiling that the full \texttt{all} shape attains.

\textbf{Some questions need several grains at once.} Interruption timing depends on evidence at three scales together, how dense the recent activity is, whether the current episode is near a boundary, and how loaded the day already is, so no single layer suffices and only the full hierarchy reaches a usable level (0.60, against 0.52 or below for any single grain). Help selection behaves the same way: the episode description alone identifies the kind of help that fits (0.82), and the full hierarchy improves on it (0.85) by adding the finer and coarser context that separates the harder cases.

\subsection{Why one universal summary is not enough}

\textbf{More content does not close the gap.} The baseline already includes the persona's role, the time of day, and the artifact currently open, yet it scores near the floor on every behavior-heavy question: the identity of the open file says nothing about whether the day has so far been deep work or light, or whether now is a good moment to interrupt. Even the \texttt{oracle} shape, which adds the declared workstream identity and workflow goal that telemetry cannot observe, never beats the best behavioral layer on any capability. Where the declared goal seems most relevant, as in choosing which help a freshly opened task needs, the observable behavioral layers already identify the task and answer it better; and where a question turns on how behavior is unfolding, the declared goal can mislead, dropping interruption timing below the baseline (0.21, the lowest cell in the study) as the model leans on a prior over the workflow in place of behavioral evidence. The strictly larger \texttt{all} shape inherits this, matching or trailing the hierarchy on six of the seven capabilities. Behavioral questions need behavioral evidence at the right grain; content, however privileged, reveals topic, not pattern.

\textbf{The levels must be addressable, not concatenated.} The lesson is not that more context helps but that the levels must stay separately \emph{addressable}: a query-conditioned agent should retrieve the motif, episode, rhythm, or combination a question requires, rather than receive every layer at once. The best-performing level shifts even \emph{within} a capability, across its sub-classes (Appendix~\ref{app:perclass}).

\section{Discussion}

\paragraph{Interpretation is scale-selective retrieval, not summarization.}
Our forecasting study and controlled ablation point the same way: behavioral structure carries real signal, but added scale is not uniformly useful, and different questions are answered at different grains, so no fixed context wins everywhere and concatenating every level can plant a misleading prior. Because the level that carries a question's answer shifts between questions, and even between the sub-classes of one question (Appendix~\ref{app:perclass}), behavioral state should be stored as addressable grains an agent retrieves by time scale rather than topic, so interpreting a long human trace becomes a retrieval problem, not only a summarization one.

\paragraph{The human side of agent-behavior interpretation.}
Interpreting agent behavior has centered on the agent's own trajectory; the complementary problem is the human trajectory it unfolds alongside. The same multi-resolution reading applies there: although our telemetry spans the whole workday rather than isolated interaction sessions, the levels transfer directly to interaction traces, where an agent could read behavior at the right grain before, during, and after it acts, to see whether its action changed what the user did next.

\paragraph{Representation, not intervention policy.}
Forecasting estimates $p(\text{future behavior}\mid\text{history})$ from telemetry; deciding whether to act asks something different, which intervention is worth its cost given the user's latent need, the price of interrupting, consent, and the counterfactual. Our representation supplies state features to such a policy but is not the policy, and the predictive strata make the same point: the episode increment is positive on average but negative for same-label revisits (Appendix~\ref{app:forecastsensitivity}), so added scale should be gated by the decision, not always retrieved.

\paragraph{Toward query-conditioned interpretation architectures.}
The failure of a stack-everything default is the concrete argument for scale-selective access. The open problems we find most pressing are learning a query-to-grain router from data, testing transfer beyond a single productivity suite, and evaluating the action-facing questions with people rather than the offline probes that bound them here.

\section{Conclusion}

A long trace of human activity has no single correct interpretation: the same events read as a brief routine at one resolution, a block of work at another, and the shape of a day at a third, and which reading an agent needs depends on the question it asks. We made this concrete with a multi-resolution vocabulary of operators, motifs, episodes, and rhythms, discovered without labels from hundreds of millions of workplace events and shown on real telemetry to be stable across independent samples, predictive of what a user does next, and never answered by a single grain across agent-facing questions. Behavioral context should therefore be kept not as one flattened stream or one summary but as separately addressable grains an agent can query by time scale. Interpreting agent behavior has largely meant making sense of what the agent did; making sense of the human trace it acts on, at the right resolution, is the complementary half, and where a genuinely helpful agent has to begin.

\section{Limitations and responsible use}

The corpus is one commercial productivity suite, so it may not carry over to other workplaces or roles, and the clusters summarize recurring patterns rather than productivity, intent, or cause. The forecasting test uses one target and a linear probe and the ablation uses synthetic fixtures and hand-picked moments, so both bound usefulness rather than settle it, and interruption timing and assistance preference still need human evaluation. We do not release the proprietary corpus or trained artifacts, though the appendices give the pipeline, settings, and compute for reproduction. Such telemetry is sensitive, so any deployment needs purpose limitation, data minimization, transparency, user control, and organizational safeguards.

\bibliographystyle{plainnat}
\bibliography{references}

\appendix
\section{Why next-episode is the forecasting target}
\label{app:target}

Behavior can be forecast at many granularities: the next action, the next motif, the next episode, or the next day-rhythm. Testing several at once, across several models and metrics, would turn a validity check into a forecasting benchmark, with each target demanding its own justification. We commit to a single target, the type of the user's next episode, because it is the only granularity that genuinely exercises a multi-scale interpretation. The next \emph{action} is too low-level: it mostly rewards instrumentation regularity and app-specific action grammar, such as \texttt{Edit} followed by \texttt{Save}, so raw operators or motifs would dominate and the coarser levels would add almost nothing. The next \emph{motif} is too close to how motifs are built, since a motif is itself an operator $n$-gram; predicting it reduces to guessing one $n$-gram from the preceding ones and says little about the levels. The next \emph{day-rhythm} is both too coarse and leaks the future, because a day's rhythm is only defined once the whole day is observed, so supplying it mid-morning would feed in behavior that has not yet happened. The next \emph{episode} sits in the useful middle: it is a coherent unit of activity, the natural granularity at which an agent reasons about where a user is in their work; it carries an objective label that needs no manual annotation; and predicting it can draw on every level at once, from recent operators and motifs to the current episode and the day so far. That is precisely where a multi-scale interpretation should matter most.

\section{Synthetic test-fixture construction}
\label{app:synthetic}

The real productivity-suite telemetry gives us the bottom of the stack, raw events, but not a ground-truth workstream, workflow, or episode label at every moment, which the controlled ablation needs. Synthetic fixtures are the right substrate for a second reason too: the behavioral representation is content-agnostic, reading only the sequence of operators and their timing and never the text of a document, message, or prompt, so a fixture that reproduces the behavioral skeleton reproduces everything the representation reads. We therefore build top-down synthetic timelines whose latent labels are known by construction and whose surface statistics are calibrated to the real data. Every LLM-driven stage below uses GPT-5.4. The cascade has six stages: (0) load and classify source task material from OdysseyBench \citep{wang2025odysseybench}; (1) extract candidate workflows from the source chat histories; (2) group workflows under business objectives into workstreams; (3) plan each workflow into behavioral episode specs; (4) match each planned episode to the top real episode clusters (of the 25 mined types, excluding noise); and (5) synthesize each episode into steps and then into timestamped action sequences over the canonical operator vocabulary. Stage 5 is where gap distributions, motif vocabulary, and app-conditioned action mix are drawn to match the bottom-up structure mined from the real event stream.

We score the fixtures with GPT-5.4 acting as an LLM judge in a separate pass: workflow and episode coherence 4.5/5 (4.49 and 4.45), behavioral fidelity against real cluster patterns 3.3/5, and inter-action gap distributions within Jensen--Shannon divergence 0.26 of the real gap model. The point is plausibility with known labels, not fidelity to any specific person.

We also check that each synthesized workflow is actually executable, both that its goal can be completed and that its latent step sequence is a valid path to that goal. For every workflow with input artifacts we programmatically materialize a testbed: input files (spreadsheets, documents, slides, PDFs, and calendar invites) that embed hidden ground-truth facts such as specific dates, names, and figures, together with deterministic checks derived from those facts, namely whether the right output file exists, the right message was sent, and the right calendar event was created. A GPT-5.4 agent with a fixed catalog of office tools (file readers and editors, spreadsheet operations, email, calendar, and a calculator) then attempts the workflow in two modes. In \emph{free} mode it sees only the workflow goal and known metadata (expected output filenames, the people involved, message subjects), which tests whether the workflow is executable in principle. In \emph{guided} mode it instead walks the pipeline's synthesized step decomposition, which tests whether that latent step sequence is itself a valid route to the goal. Each attempt is scored on three levels: which canonical operators the agent emitted, whether the required artifacts, messages, and events were produced, and an LLM judge that grades each hidden truth and an overall completeness score from 1 to 10; a workflow counts as executable when completeness reaches 7. Single-shot, this yields 85\% executable in free mode and 80\% in guided mode, so both the goals and the synthesized step sequences are task-valid.

Each moment is rendered into eight context shapes (Table~\ref{tab:shapes}); the seven capabilities and their sub-classes, with per-class sample counts, are in Table~\ref{tab:capabilities}. Questions are sampled stratified across 11 persona timelines spanning the typology range and biased toward diagnostic moments, those where the correct answer hinges on a specific grain or conjunction of grains, so the comparison between shapes is sharp. The total is 640 questions $\times$ 8 shapes $\times$ 3 runs $=$ 15{,}360 evaluations.

\begin{table}[h]
\centering
\small
\caption{The eight context shapes. The six observable shapes add layers a deployed agent could plausibly compute; \texttt{oracle} and \texttt{all} add declared metadata as reference points.}
\begin{tabular}{@{}p{0.16\linewidth}p{0.78\linewidth}@{}}
\toprule
Shape & Content on top of the common baseline \\
\midrule
baseline & persona role, time of day, the artifact under the cursor, today's calendar, recent collaborators \\
+raw & the last 60 minutes of raw events (timestamp, app, action, operation) \\
+motif & a step-level trace: per step, its goal/purpose/app and a run-compressed action sequence \\
+episode & the current and recent episode labels, durations, and boundaries \\
+rhythm & the day-so-far summary: cumulative active time, episode-type mixture, meeting and AI-assistant share \\
+hierarchy & baseline + motif + episode + rhythm together (the bottom-up pipeline's output) \\
+oracle & baseline + declared workstream identity and workflow goal (not observable from telemetry) \\
all & baseline + hierarchy + oracle combined \\
\bottomrule
\end{tabular}
\label{tab:shapes}
\end{table}

\begin{table}[h]
\centering
\small
\caption{The seven capabilities, their families, and sub-class sample counts (before the $\times 8$ shapes $\times 3$ runs).}
\begin{tabular}{@{}llp{0.40\linewidth}r@{}}
\toprule
Family & Capability & Sub-classes (sample counts) & $n$ \\
\midrule
Perception & Motif-scale recognition & drafting 33, revising 33, investigating 33, wrap-up 33 & 132 \\
Perception & Episode-scale recognition & create 35, communicate 35, consume 35 & 105 \\
Perception & Day-scale recognition & full workday 33, deep work 33, light day 26 & 92 \\
Inference & Behavior projection & same-task 25, transition 25 & 50 \\
Inference & Anomaly detection & normal 25, fragmented 25 & 50 \\
Action & Interruption timing & ideal 25, mild 25, high 25, prohibitive 22 & 97 \\
Action & Help selection & drafting 25, polish 25, analysis 23, entry 16, scheduling 25 & 114 \\
\bottomrule
\end{tabular}
\label{tab:capabilities}
\end{table}

Table~\ref{tab:questions} gives each probe verbatim, together with the answer classes that define its ground truth; the short labels match the rows of Figure~\ref{fig:scorecard} (and the capabilities in Table~\ref{tab:capabilities}). Because every moment is synthesized top-down with its latent labels fixed by construction (workstream, workflow, episode, and step), the correct answer to each question is known in advance rather than hand-annotated; the model must recover it from whichever context shape it is shown, and macro-F1 measures how often it does.

\begin{table}[h]
\centering
\small
\caption{The seven probe questions, verbatim, with the answer classes that define ground truth. Short labels match the rows of Figure~\ref{fig:scorecard}; per-class sample counts are in Table~\ref{tab:capabilities}. Each question is posed as a single multiple-choice item (with an added ``none of the above'' escape option), and every context shape answers the identical question.}
\begin{tabular}{@{}l p{0.50\linewidth} p{0.26\linewidth}@{}}
\toprule
Probe (Fig.~\ref{fig:scorecard} row) & Question, verbatim & Answer classes (ground truth) \\
\midrule
\multicolumn{3}{@{}l}{\emph{Perception}} \\
motif recog. & ``Looking at the user's recent activity (the last hour or so), which best describes the kind of work they have been doing?'' & drafting, revising, investigating, wrap-up \\
episode recog. & ``What is the user's current activity primarily aimed at?'' & create, communicate, consume \\
day recog. & ``Looking at the user's full day so far (when they started, what kinds of work, how much overall activity), which best describes the day's overall rhythm?'' & full workday, deep work, light day \\
\addlinespace
\multicolumn{3}{@{}l}{\emph{Inference}} \\
short-horizon proj. & ``Based on the user's recent activity and the patterns of their day so far, what is the user most likely to do in the next 30 minutes?'' & same-task continues, task transition \\
anomaly detect. & ``Looking at the user's recent activity (and the day so far), what kind of abnormal behavior pattern, if any, are they showing?'' & normal progress, attention fragmented \\
\addlinespace
\multicolumn{3}{@{}l}{\emph{Action}} \\
interruption timing & ``On a 1-4 scale, how disruptive would it be to surface a non-urgent assistive suggestion to the user right now (1 = ideal moment, 4 = prohibitively disruptive), considering both what the user is doing right now and the overall pattern of their day so far?'' & 1 ideal, 2 mild, 3 high, 4 prohibitive \\
help selection & ``The user has just started a new task. If the agent were to proactively offer to help take over or automate that task, which type of help would be most directly useful?'' & drafting, polish, data analysis, data entry, scheduling \\
\bottomrule
\end{tabular}
\label{tab:questions}
\end{table}

\section{Inside each family: per-class analysis}
\label{app:perclass}

\begin{figure}[t]
    \centering
    \includegraphics[width=\linewidth]{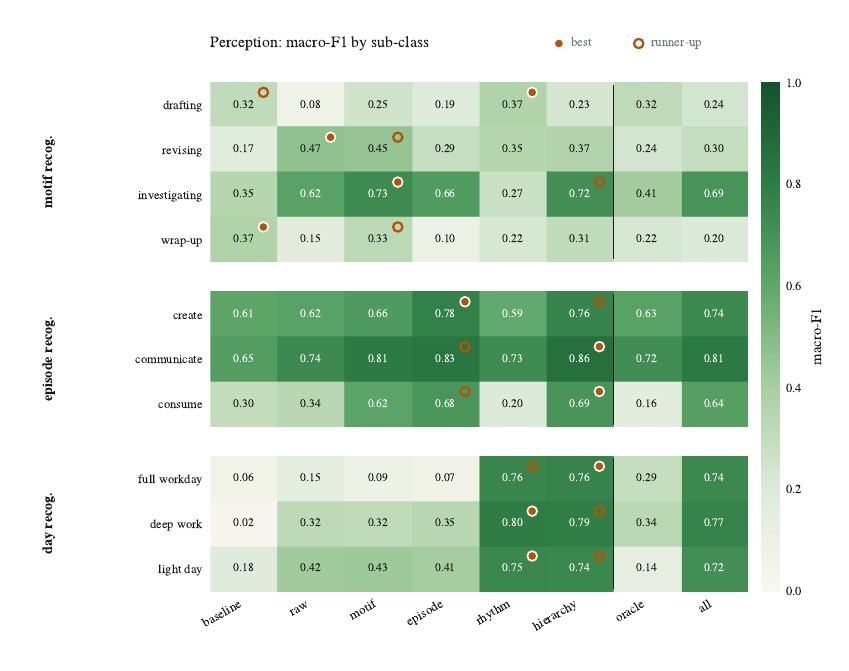}
    \caption{Perception, per sub-class. Rows are grouped by capability (motif, episode, and day recognition); a filled dot marks each row's best observable context and an open dot its runner-up. Day recognition is uniformly best at the rhythm layer, while motif recognition spreads its best cell across baseline, raw, motif, and rhythm.}
    \label{fig:pc-perception}
\end{figure}

\begin{figure}[t]
    \centering
    \includegraphics[width=\linewidth]{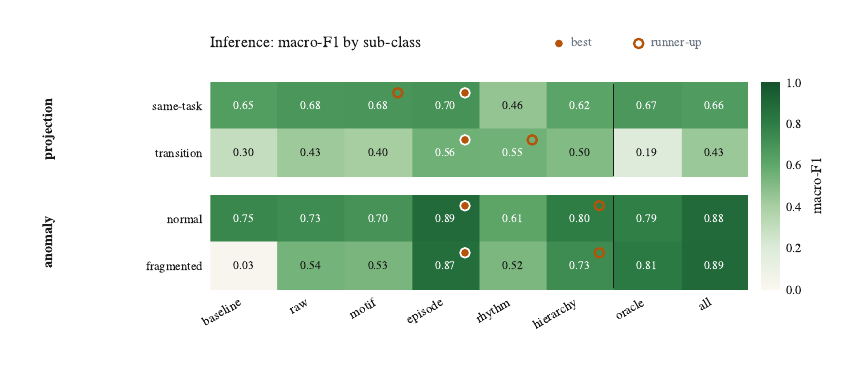}
    \caption{Inference, per sub-class. Both projection sub-classes and both anomaly sub-classes peak at the episode grain; for anomaly detection the observable episode cell already matches the \texttt{all} ceiling, so declared \texttt{oracle} metadata adds nothing.}
    \label{fig:pc-inference}
\end{figure}

\begin{figure}[hb]
    \centering
    \includegraphics[width=\linewidth]{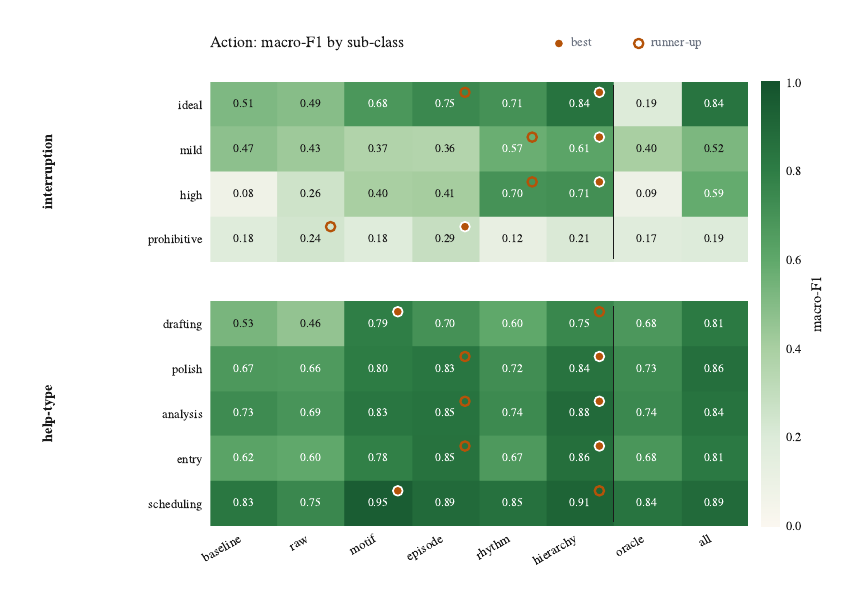}
    \caption{Action, per sub-class. Interruption timing (ideal, mild, high, prohibitive) is a broad hierarchy lead that never dominates a single row; help selection splits its best cell between motif (drafting, scheduling) and the hierarchy (analysis, entry, polish).}
    \label{fig:pc-action}
\end{figure}

The macro scorecard collapses each capability to a single F1. Splitting it back into per-class F1 makes the same point at higher resolution: the best cell moves not just between capabilities but \emph{within} a capability, across its sub-classes. Because an agent cannot know which sub-class applies until it has read the evidence, the grains a question might need must be available from the start, which is what the addressable hierarchy provides. Figures~\ref{fig:pc-perception} through~\ref{fig:pc-action} break this down one family at a time; within each figure the sub-class rows are grouped by their parent capability.

\paragraph{Perception.}
Day-scale recognition is the cleanest case: all three sub-classes score near the floor from baseline and climb sharply once the day rhythm is available. Motif-scale recognition, by contrast, is spread across several layers. Investigating is best read at the motif layer and revising at the raw-event layer, where the repeated read-edit cycle is visible before aggregation smooths it away; drafting is best read at the rhythm layer, since early-day creation tracks the shape of the day; and wrap-up is weak at every layer, because it looks different each time. No single layer is sufficient for the capability.

\paragraph{Inference.}
Anomaly detection's two sub-classes are both best read at the episode grain, attention-fragmented and normal-progress alike, already matching the \texttt{all} ceiling, so the declared workstream titles in \texttt{oracle} add nothing the observable episode cell does not already carry. Behavior projection is the family's exception: both sub-classes peak at the recent grain (episode), and the stacked hierarchy, pulled toward a misleading day-rhythm prior, underperforms the episode layer. A stack-everything default only works when every added layer is at least non-misleading.

\paragraph{Action.}
Interruption timing's lift is small but broad: the hierarchy takes the ideal-moment class outright and the contested middle (mild and high) narrowly over rhythm, while the prohibitive extreme is hard for every shape (the model is reluctant to commit at that end). Help selection is more concentrated: every sub-class peaks at an abstracted layer, because the episode description and step-level motif together name the task the help should target. Motif is best on the two cleaner classes (drafting and scheduling) and the hierarchy on the harder ones (analysis, entry, and polish).

\newpage
\section{Predictive-validity sensitivity analyses}
\label{app:forecastsensitivity}

The episode increment (S-C3 minus S-C2, the pre-registered primary) is positive overall but not uniform across strata (Table~\ref{tab:forecastsens}). It is largest for transitions into a new episode type (state changes, $+0.035$, or $+20\%$ relative to that stratum's S-C2 score) and for near-immediate next episodes (a 0--10\,s gap, $+0.032$), and it is \emph{negative} for same-label revisits ($-0.098$, or $-27\%$), where immediate operator context already suffices and coarser structure adds noise. The increment is stable across the five user-persona strata ($+0.017$ to $+0.025$) and essentially unchanged when always-on service accounts are excluded ($+0.022$). These strata are descriptive and were not part of the pre-registered primary comparison. The negative same-label-revisit stratum is the predictive-side evidence that added scale should be gated rather than always retrieved.

\begin{table}[h]
\centering
\small
\caption{Episode-increment (S-C3$-$S-C2) macro-F1 differences by stratum, held-out test set. The increment is also stable across the five user-persona strata ($+0.017$ to $+0.025$), omitted here for space. Descriptive; not part of the pre-registered primary comparison.}
\begin{tabular}{@{}llrr@{}}
\toprule
Stratum & Group & Episode increment & Transitions \\
\midrule
Overall & all & $+0.023$ & 2{,}975{,}438 \\
Transition kind & state change & $+0.035$ & 2{,}570{,}957 \\
Transition kind & same-label revisit & $-0.098$ & 404{,}481 \\
Future gap & 0--10\,s & $+0.032$ & 363{,}134 \\
Future gap & $\geq$300\,s & $+0.020$ & 2{,}612{,}304 \\
Always-on & excluded & $+0.022$ & 2{,}967{,}600 \\
\bottomrule
\end{tabular}
\label{tab:forecastsens}
\end{table}

\section{Structural-validity figure}
\label{app:stability}

\begin{figure}[h]
    \centering
    \includegraphics[width=\linewidth]{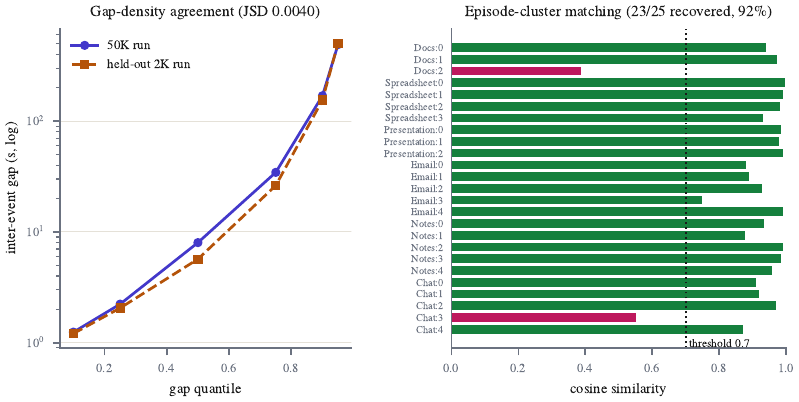}
    \caption{Independent held-out replication. \textbf{Left:} log-gap quantiles for the 50K run overlap those of a disjoint 2K run (Jensen--Shannon divergence \stabilityGapJSD). \textbf{Right:} per-application episode clusters matched by Hungarian assignment; \stabilityEpisodeRecovery{} of 25 types are recovered above the $0.7$ cosine threshold.}
    \label{fig:stability}
\end{figure}

Figure~\ref{fig:stability} shows the independent held-out replication behind our structural-validity result: the log-gap density overlay and the per-application episode-cluster matches between the disjoint 2{,}000-user re-run and the full 50{,}000-user run.

\section{Day rhythms and personas}
\label{app:rhythms}

\begin{figure}[h]
    \centering
    \includegraphics[width=\linewidth]{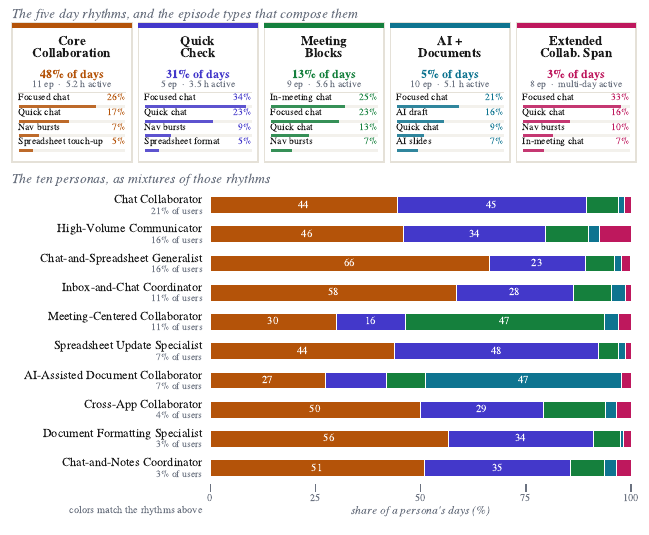}
    \caption{The five day-rhythm archetypes over 1.82M user-days, ordered by prevalence: Core Collaboration (48\%), Quick-Check (31\%), Meeting-Block (13\%), AI-Assisted Document (5\%), and Extended Collaboration Span (3\%). Each card shows the rhythm's share of user-days, a short profile, and the episode types that compose it, and doubles as the color key. \textbf{Bottom:} the ten personas as \emph{dialects}, each a longitudinal mixture over the same five rhythms. All names are generated from behavioral profiles after clustering and are never inputs to the objective.}
    \label{fig:dayrhythms}
\end{figure}

Figure~\ref{fig:dayrhythms} shows the five day-rhythm archetypes and the ten personas discovered over the full corpus. They demonstrate that a single multi-resolution vocabulary describes a heterogeneous population, and are not a taxonomy the paper's claims depend on.

\section{Compute resources}
\label{app:compute}

The reported experiments use no local accelerators. The dominant cost is LLM inference through a hosted GPT-5.4 deployment, and the call volume is modest: the two-pass operator canonicalization labels roughly 14{,}000 distinct $(\text{app},\text{action})$ pairs rather than individual events; the synthetic fixtures are generated and then rated by an LLM judge once each; and the resolution ablation is $640\times8\times3=15{,}360$ answering calls. All non-LLM computation, the Gaussian-mixture gap segmentation, per-application $k$-means, truncated SVD, Hungarian matching, and the L2-regularized SGD probes, runs on a single commodity multi-core CPU workstation and is inexpensive relative to LLM inference; the one-time bottom-up mining over the full corpus is the largest of these steps. The overall project used more compute than the final runs alone, owing to pipeline iteration and smaller-scale pilot ablations that are not reported here.

\section{Assets and licenses}
\label{app:licenses}

We use OdysseyBench \citep{wang2025odysseybench} as the source task material for the synthetic fixtures (Appendix~\ref{app:synthetic}), under its MIT License. The LLM stages use GPT-5.4 through a hosted commercial API, under its terms of use. All classical machine learning uses scikit-learn and SciPy (both BSD-3-Clause) and NumPy (BSD-3-Clause). We use each asset in accordance with its license or terms.

% \clearpage
% \input{checklist.tex}

\end{document}